\documentclass[sigconf]{acmart}

\usepackage{tabularx}
\usepackage{svg}
\usepackage{multirow}
\usepackage{booktabs}
\newcolumntype{L}{>{\centering\arraybackslash}m{3cm}}
\usepackage{tabularray}
\usepackage{array}
\usepackage{diagbox}

\usepackage{fmtcount}

\usepackage{caption}

\usepackage{algorithm}
\usepackage{algpseudocode}

\usepackage{svg}

\AtBeginDocument{%
  \providecommand\BibTeX{{%
    \normalfont B\kern-0.5em{\scshape i\kern-0.25em b}\kern-0.8em\TeX}}}

\setcopyright{acmcopyright}
\copyrightyear{2023}
\acmYear{2023}
\acmDOI{XXXXXXX.XXXXXXX}
\setcopyright{acmcopyright}
\copyrightyear{2018}
\acmYear{2018}
\acmDOI{XXXXXXX.XXXXXXX}

\acmConference[Arxiv]{2023}{2023}{}

\begin{document}
\title{KITLM: Domain-Specific Knowledge InTegration into Language Models for Question Answering}

\author{Ankush Agarwal}
\authornote{Both authors contributed equally to this research.}
\email{ankushagrawal@cse.iitb.ac.in}
\author{Sakharam Gawade}
\authornotemark[1]
\email{sakharamg@cse.iitb.ac.in}
\affiliation{%
  \institution{IIT Bombay}
  \country{India}
}

\author{Amar Prakash Azad}
\email{amarazad@in.ibm.com}
\affiliation{%
  \institution{IBM Research}
  \country{India}
}
\author{Pushpak Bhattacharyya}
\email{pb@cse.iitb.ac.in}
\affiliation{%
  \institution{IIT Bombay}
  \country{India}
}

\renewcommand{\shortauthors}{Ankush and Sakharam, et al.}

\begin{abstract}

Large language models (LLMs) have demonstrated remarkable performance in a wide range of natural language tasks. However, as these models continue to grow in size, they face significant challenges in terms of computational costs. Additionally, LLMs often lack efficient domain-specific understanding, which is particularly crucial in specialized fields such as aviation and healthcare. 
  To boost the domain-specific understanding, we propose, \emph{KITLM}\footnote{The URL for our dataset and source codes is:\\ \url{https://github.com/sakharamg/KITLM}}, a novel knowledge base  integration approach into language model through relevant information infusion.  By integrating pertinent knowledge, not only the performance of the language model is greatly enhanced, but the model size requirement is also significantly reduced while achieving comparable performance. 
Our proposed knowledge-infused model surpasses the performance of both GPT-3.5-turbo and the state-of-the-art knowledge infusion method, SKILL, achieving over $1.5$ times improvement in exact match scores on the MetaQA. 
KITLM showed a similar performance boost in the aviation domain with AeroQA.  
The drastic performance improvement of KITLM over the existing methods can be attributed to the infusion of relevant knowledge while mitigating noise.  
In addition, we release two curated datasets to accelerate knowledge infusion research in specialized fields: a) AeroQA, a new benchmark dataset designed for multi-hop question-answering within the aviation domain, and b) Aviation Corpus, a dataset constructed from unstructured text extracted from the National Transportation Safety Board reports. 
Our research contributes to advancing the field of domain-specific language understanding and showcases the potential of knowledge infusion techniques in improving the performance of language models on question-answering.

\end{abstract}





\maketitle

\section{Introduction}
Large pre-trained language models (PLMs) \cite{raffel2020exploring} have succeeded remarkably in various NLP downstream tasks. 
Their achievements can be attributed to two key factors: extensive pre-training on diverse text sources and the ability to fine-tune domain-specific data.
PLMs undergo
extensive pre-training on vast amounts of text data from various sources such as books, articles, and websites. This process allows them to develop a profound understanding of language and capture a comprehensive range of linguistic patterns and contextual information.
Furthermore, PLMs can be fine-tuned on domain-specific datasets, enabling them to specialize and adapt to a particular domain. This fine-tuning process refines the models' knowledge and performance, allowing them to excel in tasks specific to those domains.
However, recent research has highlighted the efficacy of incorporating knowledge graphs into language models using diverse techniques \cite{saxena-etal-2022-sequence, moiseev-etal-2022-skill, zhang2022greaselm, yasunaga2021qa}.
Our paper shows that incorporating relevant structured knowledge from knowledge graphs can further enhance language model performance and domain-specific understanding.
\par
A knowledge graph (KG) is a graph-based structure comprising real-world entities represented as nodes, such as Ginger Rogers and Primrose Path, and relationships between them represented as edges, such as Ginger Rogers | starred\_actors | Primrose Path. KGs can be specific to a particular domain \cite{agarwal-etal-2022-knowledge} or general in nature \cite{vrandevcic2014wikidata}. These knowledge graphs, which serve as knowledge bases, play a vital role in knowledge-intensive applications like question answering, as they provide structured and organized information for effective retrieval and analysis.
\par
Various studies have explored different methods for infusing knowledge into language models. One popular approach involves verbalizing triples in the knowledge base and continually pretrain the LLM using 
a training criteria 
such as masked language modeling. However, this approach can be computationally demanding. Other methods like QA-GNN \cite{yasunaga2021qa} and GreaseLM \cite{zhang2022greaselm} rely on knowledge graph embeddings \cite{Knowledge_Graph_Embeddings_Survey_Dai} to obtain the domain knowledge which requires additional training. 
The two critical factors in a knowledge infusion method are: i) the quality of infused knowledge, which allows for achieving strong empirical performance, and ii) the simplicity of the architecture. These underscore the need for a knowledge infusion technique that is computationally efficient while maintaining high quality and simplicity.
\par
Our paper presents an innovative framework for integrating knowledge into language models like T5 \cite{2020t5} through fine-tuning. The experimental results demonstrate that the checkpoints trained using the proposed approach on AviationKG \cite{agarwal-etal-2022-knowledge} and WikiMovies \cite{miller-etal-2016-key} outperforms the T5 baselines, state-of-the-art SKILL \cite{moiseev-etal-2022-skill} and GPT-3.5-turbo on MetaQA \cite{zhang2018variational} and our curated multihop QA dataset, AeroQA.
Instead of introducing additional parameters to pre-trained language models (PLMs) or modifying their architectures, the proposed framework employs a novel knowledge integration objective. This objective entails verbalizing the KG triples, extracting pertinent triples for each question-answer pair using ColBERTv2 \cite{santhanam-etal-2022-colbertv2}, and incorporating them during both the training and testing phases of the language model.
\par
Language models have impressive capabilities in acquiring knowledge from unstructured text corpora. However, their ability to learn and retain new information directly from structured knowledge graphs or their associated textual descriptions is still uncertain.
Structured knowledge graphs provide explicit relationships and connections between entities, enabling a more organized representation of knowledge than unstructured text. However, incorporating this structured information into LMs poses challenges due to their inherent architecture, primarily trained on predicting the next word given the context.
The extent to which LMs can effectively learn and retain new information directly from structured knowledge graphs or accompanying text descriptions is an area of active exploration and research. Conducting research is necessary to make further advancements and fully exploit the potential of LMs in effectively integrating structured knowledge into their learning processes.
\par
We conducted a comprehensive study to enhance our proposed framework, KITLM, by exploring the impact of different formats of external knowledge on a language model. We incorporated unstructured general corpora, domain-specific corpora, and structured knowledge (triples) into the T5 model for question-answering. To evaluate the effectiveness of these settings, we employed the SKILL approach \cite{moiseev-etal-2022-skill} and compared T5, T5 + unstructured text, T5 + KG triples, and T5 + unstructured text + KG triples on the AeroQA and MetaQA datasets. Our proposed approach, KITLM, outperformed all other settings, underscoring the importance of integrating relevant knowledge alongside LLMs while mitigating noise for enhancing question-answering capabilities.
\par
Our contributions are:
\begin{enumerate}


    \item Introduce two datasets to accelerate knowledge infusion research in specialized fields: (a) AeroQA, a closed-book question-answering dataset with multi-hop reasoning. It contains 34k QA pairs, with 21k 1-hop pairs and the rest being 2-hop pairs. (b) Aviation Corpus, comprising 665,000 lines of clean English text from 4,000 NTSB \footnote{\url{https://www.ntsb.gov/Pages/AviationQuery.aspx}} reports. It is specifically curated for continual pre-training to facilitate knowledge infusion tasks in language models.
    


    \item KITLM, a novel framework introduces a seamless integration of relevant verbalized triples from a knowledge base into the language model without modifying its architecture. Leveraging ColBERTv2, KITLM extracts the most pertinent triples associated with each instance in the question-answering dataset. 
   Our approach surpasses the state-of-the-art knowledge infusion method, SKILL, by more than 20\% on both AeroQA and MetaQA datasets.
   



\item KITLM > GPT-3.5-turbo; Our knowledge-infused model surpasses GPT-3.5-turbo by over 1.5 times in AeroQA and MetaQA, highlighting the significant reduction in the requirement of language model size through relevant knowledge infusion.


    
    
\end{enumerate}

\section{Motivation}
Aviation-related datasets are scarce and highly sought after, posing challenges for building question-answering (QA) systems capable of reasoning over knowledge graphs like AviationKG \cite{agarwal-etal-2022-knowledge}. To address this, we have developed a valuable multi-hop reasoning QA dataset derived from the National Transportation Safety Board reports in the aviation domain. This dataset is valuable for the aviation industry and researchers, facilitating information retrieval and QA tasks. Its creation aims to provide deeper insights into aircraft accidents and contribute to developing preventive measures to enhance aviation safety.
\par
Large Language Models \cite{brown2020language, scao2022bloom} have demonstrated efficient performance across various downstream NLP tasks. However, the high computational requirements associated with LLMs have raised concerns. Furthermore, LLMs are typically trained on generic datasets, so their suitability for domain-specific tasks is limited. Our study provides evidence that computational resources can be conserved by employing smaller language models for specific tasks. Additionally, we highlight the importance of integrating relevant knowledge in the LM for addressing the needs of domain-specific tasks.

\section{Background and Related Work}


\begin{figure*}
  \centering
  \includegraphics[width=0.9\textwidth]{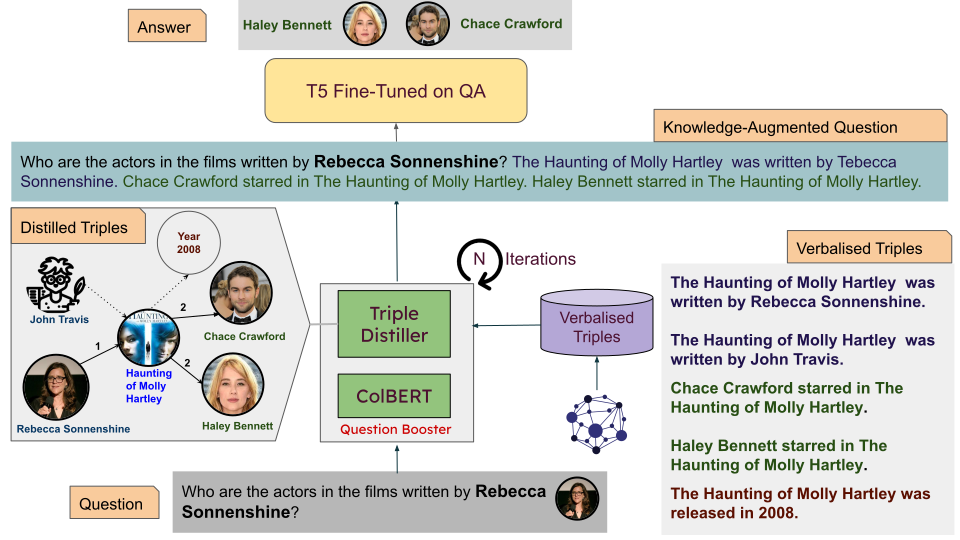}
  \caption{The proposed framework, KITLM, is illustrated in the flow diagram. Initially, triples are extracted from knowledge bases like the WikiMovies dataset \cite{miller-etal-2016-key} and transformed into verbalized form. Subsequently, ColBERTv2 \cite{santhanam-etal-2022-colbertv2} is employed to retrieve the top-K relevant triples related to the given question from the set of verbalized triples. The triples are distilled N times for the N-hop question-answering. The distilled triples are then concatenated with the question and provided as input to the fine-tuned T5 to generate an answer.}
  \label{fig:approach}
\end{figure*}


Prevalent state-of-the-art models like BERT \cite{devlin-etal-2019-bert}, GPT-3 \cite{brown2020language}, and T5 \cite{raffel2020exploring} have emerged as powerful tools for various tasks. These models are typically pre-trained on unstructured text data, allowing them to comprehend language within a contextual framework. However, knowledge about the real world is crucial to gain a comprehensive understanding of a statement. This world knowledge is frequently represented as triples within a knowledge graph.
\\\\
\textbf{Knowledge Graph Question Answering.} A Knowledge Graph (KG) is a collection of entities and their relationships, represented as triples (subject, relation, object). 
KGs are commonly stored in a triple format, ranging from large-scale KGs like Wikidata \cite{vrandevcic2014wikidata} to small-scale KGs such as those in \citep{miller-etal-2016-key} and \citep{agarwal-etal-2022-knowledge}. 
KGs are particularly valuable when accurate information can be extracted from them. Initially, querying KGs in Natural Language (NL) involved rule-based \cite{guo2020survey} and pattern-based systems \cite{affolter2019comparative}. 
Semantic parsing \cite{bast2015more} was also utilized for solving these queries by converting NL questions into symbolic queries over the KG. 
However, recent advancements have shifted towards the adoption of sequence-to-sequence (seq2seq) architectures \cite{zhong2017seq2sql} and pre-trained models, harnessing the power of neural networks.
\\\\
\textbf{Knowledge infusion.} Extensive research on querying knowledge graphs in natural language has driven the development of diverse methods for knowledge retrieval, addressing the challenge of converting natural language into graph query language. A particularly successful approach involves combining knowledge graphs with deep learning (DL), which has generated considerable interest among researchers due to the increasing significance of knowledge globally. 
One commonly used approach for incorporating structured knowledge into models is to convert the knowledge into natural language text. ERNIE 3.0 \cite{sun2021ernie} adopts this approach by training a knowledge-enhanced model on a corpus that combines triples and their corresponding sentences. During training, random masking is applied to either the relation in a triple or words in a sentence.
Methods like QA-GNN \cite{yasunaga2021qa} and GreaseLM \cite{zhang2022greaselm} employ knowledge infusion techniques that involve propagating information through a graph to capture the dependencies and relationships among entities. 
\par 
The synergy of KG and DL can be categorized into two groups: 
a) Utilizing KGs during inference, as demonstrated in studies like PullNet \cite{sun-etal-2019-pullnet}.
b) Infusing knowledge into model weights during pre-training, as explored in approaches such as K-BERT \cite{Liu2020KBERTEL}, KGT5 \cite{saxena-etal-2022-sequence} and SKILL \cite{moiseev-etal-2022-skill}.
This paper examines the SKILL technique for infusing knowledge into language models (LMs) during pre-training, and a novel framework called KITLM is introduced for knowledge infusion during inference in LMs. KITLM uniquely incorporates relevant knowledge into language models while effectively mitigating noise, a feature lacking in previous infusion methods.

\section{Methodology}

This section details the following methodologies for knowledge integration:
\begin{enumerate}

    \item Our novel framework \emph{KITLM}, designed for multi-hop question answering, depicted in Figure \ref{fig:approach}.

        \item The T5 pre-trained model and its continual pre-training using structured knowledge, unstructured corpora, including C4 and Aviation Corpus, inspired by the state-of-the-art infusion method, SKILL \cite{moiseev-etal-2022-skill}.
        
\end{enumerate}

\subsection{Knowledge Integration for Multi-hop QA} \label{KITLM}


An overview of the KITLM framework is depicted in Figure \ref{fig:approach}. 
Knowledge infusion method, especially question booster(explained later), in KITLM is designed in a way to be more effective for multi-hop question answering. KITLM exhibits adaptability across various domains without requiring modifications to the language model's architecture, as long as a knowledge graph is available. KITLM can also be incorporated with most of the language models seamlessly independent of the LM architecture.
\par
\subsubsection{Task Formulation} KITLM obtains the answer to the question using two stages, namely retrieval and prediction stages. Let $A$ be the set of potential  answers and $a^{*}$ be the predicted answer where $a^{*} \in A$, $Q$ be the set of questions and $\textbf{q}$ be the input question, and $\theta$ represents the weights of models used in KITLM. Then the predicted answer
\begin{equation}
    a^{*} = argmax_{a \in A} P(a|\textbf{q};\theta).
\end{equation}
where $P(a|\textbf{q};\theta)$ represents the probability of answer given a question. $P(a|\textbf{q};\theta)$ 
can be decomposed into the retrieval stage and prediction stage.  

\indent \textbf{Retrieval stage}: 
Given a question q, we retrieve a set of triples $t\in T$ using an iterative retrieval mechanism, where $T$ is the set of triples used as context for the prediction stage. We can decompose $P(a|q;\theta)$ as:
\begin{equation}
P(a|q) = \sum_{t \in T}P(a|t,q;\theta_{p}) P(t|q;\theta_{r}).
\end{equation}
%

Here, $P(a|t,q;\theta_{p})$ is the probability of the answer $a$ given question $q$ and triples $t$ to the predictor and $P(t|q;\theta_{r})$ is the probability of a triple $t$ given a question $q$ to the retriever. Further, the set of triple $t$ can be rewritten as the set of relevant triples retrieved at each retrieval step helpful for answering a $K$ hop question. Let $t^k$ represent the set of triples retrieved at the $k^{th}$ hop where $t^k \in T$. It can be written as:
\begin{equation}
P(t|q) = P(t^{K},t^{K-1},...,t^{k},...,t^{2},t^{1} | q).   
\end{equation}

For retrieving $t$ at the $k^{th}$ step, we need $t^{k-1}$ ... $t^1$ along with question $q$. Therefore using chain rule, we can express $P(t|q)$ as:
\begin{equation}
    P(t|q) = \prod_{k=2}^{K} P(t^{k} |t^{k-1},...,t^{1},q) P(t^{1}|q).
\end{equation}
$P(t|q)$ can be interpreted using the score used by the retriever for ranking the triples i.e. $P(t|q) \propto S_{q,t}$. 
We use ColBERTv2 \cite{santhanam-etal-2022-colbertv2} as the retriever to score the triples as $S_{q,t} = \sum _{i=1}^{N} max_{j=1}^{M} Q_{i} T_{j}$, where each query token's representation $Q_i$ is aligned with the most relevant triple token representation $T_j$.
\par
\indent \textbf{Prediction stage:} 
To predict the answer $a^*$, we input a question \textbf{q} and its relevant triples \textbf{t} to a T5 fine-tuned model for QA task. $\textbf{q}</s>\textbf{t}$ is the input to the model and $a^*$ is the output obtained through greedy decoding. Here, $</s>$ is the separator token used by T5.
\par

\subsubsection{KITLM Algorithm}
In this section, we describe the detailed methodology of the retrieval and prediction stages implemented in KITLM which is shown in figure \ref{fig:approach}. KITLM recognizes the relevance of integrating triples with input questions as contextual information to improve question-answering accuracy. The integration process begins with the verbalization of extracted triples from a knowledge graph. To identify the most pertinent triples for a given question, ColBERTv2 is employed to index the verbalized triples. The highest-ranked triples are selected as the context for the question during the fine-tuning process.
\par
Since the integration of knowledge relies heavily on the retriever, we use ColBERTv2 because of its high performance in both in-domain and out-of-domain information retrieval. However, in case of multi-hop questions, the retrieval is likely to be highly noisy even with ColBERTv2. To alleviate this problem, we propose an iterative approach where after every retrieval iteration, we filter out the noise using the \textit{triple distiller}.
\par
 We repeat the distilling process N times for a N-hop question. Additionally, after each iteration, the distilled triples are appended to the question and used as additional input for querying ColBERTv2 in the subsequent iteration. This augments the query with additional knowledge after each iteration. The method of question augmentation through triple distiller is called \textit{Question Booster} as shown in figure \ref{fig:approach}. The iterative process is outlined in Algorithm \ref{pseudo_code}.
\noindent More formally, the retrieval process involves the following steps:
\
\begin{enumerate}
    \item Initialization: An empty set ($E^\prime$) is created, initially containing the entities mentioned in input question $Q_0$.
\item Knowledge Booster: This comprises of the following stages,
\begin{enumerate}
\item Triple Distillation: Top triples are retrieved from the knowledge base using ColBERTv2 in the first iteration. These retrieved triples are then filtered based on the presence of entities in $E^\prime$. Only the triples that include entities from $E^\prime$ are retained, while others are discarded. Additionally, we also exclude the triples visited in previous iterations. These filtering steps help in gathering relevant triples and reduce noise that may confuse the language model.

\item  Question Knowledge Augmentation: The filtered triples are merged with the input question to create an updated format ($Q_1$). 
\end{enumerate}

\end{enumerate}


\begin{algorithm}[H]
\caption{Retrieving context for N-hop Question Answering with the KITLM approach. The context in this case comprises the pertinent triples extracted from the knowledge base.}
\label{pseudo_code}
\begin{algorithmic}[1]
\Require
  \State $Q_0$ {$\rightarrow$ Input Question}
  \State $T$ {$\rightarrow$ Triples in a Knowledge Graph}
  \State $N$ {$\rightarrow$ Number of hops in N-hop Question}
  \State $E$ {$\rightarrow$ Set of Entities}
    \State ${ColBERT}(Q_i | T)$  {$\rightarrow$ ColBERTv2 indexed on T}
\Ensure
  \State $Q_{N}$ {$=$ $Q_0 + Relevant Triples$}

\Procedure{N-HopQA} {$Q_0,T,E,N$}
\State $E^{\prime} \gets entities\ in\ Q_0$
  \For{$i = 0$ to $N-1$}
    \State $Ret \gets \text{ColBERT}(Q_i | T)$ \Comment{Retrieve top triples, $Ret \subseteq T$}
    \State $Fil \gets \text{Triples}\in Ret\ \text{having}\ Entities\in E^{\prime}  $ \Comment{Clean the retrieved triples}
    \State $Q_{i+1} \gets Q_i + Fil$ \Comment{Append the filtered triples to $Q_i$}
    \State $E^{\prime} \gets \text{entities in}\ Fil \setminus E^\prime$ \Comment{New entities in the filtered triples}
  \EndFor
  \State \Return $Q_{N}$
  \EndProcedure

\end{algorithmic}
\end{algorithm}

Following these steps iteratively, the multi-hop question-answering system gradually gathers relevant triples from the knowledge base, avoids repetition, and maintains an updated entity set to guide the retrieval process. 
The iterative loop is repeated for N iterations to achieve optimal N-hop question answering (QA) results.
The retrieval process enhances the model's ability to generate accurate answers by continuously refining available information. As a result, the accuracy of multi-hop question answering improves.
\par
The integration of relevant triples with the questions in KITLM encompasses the entire QA dataset, which includes the train, validation, and test sets.
During this process, the input question is denoted as $Q_0$, while the retrieved sequence of triples is represented as $Fil$. The input provided to the language model is constructed as "question: $Q_0$</s>context: $Fil$". After the integration of triples, the language model undergoes fine-tuning using the training data. Following the fine-tuning process, the model is utilized on the test set to generate question-answering results.

\subsection{Structured Knowledge Infusion for Language Models}



In this section, we investigate incorporating knowledge into language models using knowledge triples and textual information. We delve into the details of infusing knowledge into LMs by training the T5 model on unstructured corpora, and factual triples extracted from knowledge graphs. We compare the performance of different models, namely: 
a) T5-large as the baseline model,
b) T5-large + textual information,
c) T5-large + KG triples, and
d) T5 + textual information + KG triples.
The results of these models are presented in the Table \ref{tab:KG_DL_results}.
\par
The knowledge infusion method during pre-training is inspired from SKILL.
In this approach, triples are extracted from the knowledge graph and combined with the text to prevent any degradation in the model's performance on natural language understanding tasks. For the MetaQA dataset, the C4 text is utilized, while the curated Aviation corpus (Section \ref{avi_corpus}) is employed for the AeroQA dataset (Section \ref{aeroqa}).
A subset of the C4 corpus and Aviation corpus equal to the number of triples in the KG is used for continual pre-training of the language model.
After combining the triples and text, the T5 model is continually pre-trained using a salient masked language modeling technique. 
\par
T5, a text-to-text transfer transformer model, was initially trained on the C4 corpus using a masked-language modeling technique. In this approach, certain spans of tokens in a sequence are randomly masked, and the model predicts the missing tokens. The approach described in \cite{roberts-etal-2020-much} is followed where instead of masking random tokens, salient terms are masked to improve performance on downstream tasks that require a deeper understanding of the sequence, such as question answering \cite{salient_span_pmlr-v119-guu20a}.
The salient terms are the entities found within the corpora and knowledge graphs. 
The entities in C4 with the highest predicted probability is masked by a BERT \cite{devlin-etal-2019-bert} model finetuned on the CoNLL 2003 NER dataset \footnote{\url{https://huggingface.co/dslim/bert-base-NER}} \cite{CONLL_tjong-kim-sang-de-meulder-2003-introduction}. For aviation corpus, we additionally masked NERs and nouns detected by Spacy \footnote{\url{https://spacy.io/}} since entities in AviationKG and NTSB reports can also be compound nouns. E.g. Visual Conditions is an entity in the AviationKG's triple: AccidentNumber\_LAX05LA060 | hasConditionsAtAccidentSite	| Visual Conditions. For AviationKG and Wikimovies, we randomly masked either the head entity or tail entities.
\par
By following the described continual-training process, a T5 model is transformed into a knowledge-infused model. Subsequently, the trained model is fine-tuned for the specific task, which in our case is question answering, leading to the creation of the fine-tuned model. The fine-tuned model is employed to generate answers for the test-set.




\section{Dataset}

This section provides a comprehensive explanation of the dataset creation process to facilitate research on knowledge infusion. It introduces AeroQA, a benchmark dataset specifically designed for question-answering tasks in the aviation domain. Additionally, an aviation-related text dataset called Aviation Corpus, similar to C4, is created. In Section \ref{exp_data}, detailed information is presented about the experimental data, including AviationKG and WikiMovies, along with their corresponding question-answering pairs. 

\subsection{AeroQA: A Benchmark Dataset for Aviation Domain} \label{aeroqa}

To address the limitations of the AviationQA \cite{agarwal-etal-2022-knowledge} dataset and evaluate the reasoning ability over the AviationKG knowledge graph, we have created AeroQA, a multi-hop question-answering dataset in the aviation domain. While AviationQA is a large dataset in the aviation domain, it is limited in two key aspects. Firstly, all the questions in AviationQA are single-hop, which does not allow for evaluating the model's ability to reason over knowledge graphs like AviationKG. Secondly, only a fraction of AviationQA pairs contain questions that can be answered using the triples from AviationKG, limiting the utilization of the full reasoning potential of the QA pairs. AeroQA is specifically curated to overcome these limitations and provide a dataset that facilitates reasoning over KGs in the aviation domain. 
\par
AeroQA is a multi-hop closedbook QA
dataset for the aeronautics domain.
This dataset complements the pre-processed AviationKG knowledge graph and enables reasoning tasks. The AviationKG is constructed from the National Transportation Safety Board (NTSB) reports which contain information about aircraft accidents and their investigation. AeroQA consists of a comprehensive collection of 34k questions specifically designed to assess both single-hop and multi-hop reasoning abilities.
Out of these QA pairs, 21k are 1-hop QA pairs, while the remaining are for 2-hop reasoning.
The dataset is divided into three parts: training, validation, and testing, with an 80:10:10 split ratio. To provide an overview of the dataset, Table \ref{tab:QA_stats} presents the distribution of 1-hop and 2-hop questions. Below, we present a selection of examples from the AeroQA dataset to provide a glimpse into its content.
\\\\
Examples of One-hop Questions in AeroQA:
\begin{itemize}
    \item Q: What certificate does [Pilot\_ATL03LA101] have? \\
    A: Private

    \item Q: What is the engine manufacturer associated with [Registration\_N127RB]? \\
    A: Lycoming 

    \item Q: What caused [AccidentNumber\_FTW93LA202]? \\
    A: Pre-Flight Planning | Fluid Fuel | Terrain Condition
\end{itemize}

Examples of Two-hop Questions in AeroQA:
\begin{itemize}
    \item Q: What is the aircraft category of the registered aircraft involved in [AccidentNumber\_CHI03LA242]? \\
    A: Airplane | Gyroplane

    \item What could have contributed to the cause of the accident [AccidentNumber\_SEA96TA046]? \\
    A: Pilot in Command | Pilot of other Aircraft | Check Pilot
\end{itemize}

\noindent The AeroQA dataset contains multiple answers for each question, which are separated by the `|' symbol. The entities mentioned in the questions are enclosed within square brackets `[]'. These entities are present in the AviationKG knowledge base.
The dataset consists of 87 relations for the 1-hop question-answer pairs and 35 relations for the 2-hop question-answer pairs. These relations serve as templates for constructing the question-answer pairs. The template generation process involved using the prompt-based approach with ChatGPT \cite{chatgpt}, where different relations along with their head and tail entities were used as prompt text. The model was then prompted to generate the template for the question-answer pairs. The generated output was subsequently filtered and manually checked to form the final question-answer templates.
The structure of the templates in the AeroQA dataset is exemplified in Table \ref{AeroQA_1-hop} for 1-hop questions and Table \ref{AeroQA_2-hop} for 2-hop questions.


\begin{table*}[!ht]
  \centering
  \captionsetup{position=below} 
  \setlength{\abovecaptionskip}{10pt}
  \begin{tabular}{>{\centering\arraybackslash}p{5cm} 
 >{\centering\arraybackslash}p{10cm}}
    \hline
    \textbf{Relation} & \textbf{Template} \\
    \hline
    
    hasAircraftManufacturer & What is the aircraft manufacturer associated with [HEAD] \\
    
    hasFederalAviationRegulation & What is the Federal Aviation Regulation associated with [HEAD] \\
    
    OccurredAtCountry & In which country did [HEAD] occur \\
    \hline
  \end{tabular}
  \caption{The table displays the templates employed in constructing the AeroQA 1-hop dataset. These templates utilize the placeholder [HEAD], which corresponds to the head entity of the KG triples, \textit{i.e.}, accident number, and registration number present in the NTSB report.}
  \label{AeroQA_1-hop}
\end{table*}


\begin{table*}[!ht]
  \centering
  
    \captionsetup{position=below} 
  \setlength{\abovecaptionskip}{10pt}
  
  \begin{tabular}{>{\centering\arraybackslash}p{4cm} >{\centering\arraybackslash}p{4cm} >{\centering\arraybackslash}p{7cm}}
    \hline
    \textbf{Relation1} & \textbf{Relation2} & \textbf{Template} \\
    \hline
    
    hasRegistrationNumber & hasAirworthinessCertificate & What is the airworthiness certificate of the registered aircraft involved in [HEAD] \\
    
    IsCausedBy & IsCausedDueTo & What could have contributed to the cause of the accident [HEAD] \\
    
    hasPilot & hasInstructorRating & What was the instructor rating of the pilot in the aircraft involved in [HEAD] \\
    \hline
  \end{tabular}
  \caption{The table showcases the templates used for constructing the AeroQA 2-hop dataset. In these templates, the placeholder [HEAD] represents the head entity of the KG triples, \textit{i.e.}, accident number, and registration number of the NTSB report, which is utilized to generate the 2-hop AeroQA pairs.}
  \label{AeroQA_2-hop}
\end{table*}

\subsection{Aviation Corpus: A dataset consisting of Aviation text} \label{avi_corpus}

The MetaQA dataset requires C4 \cite{raffel2020exploring} corpus for the MLM training with the SKILL \cite{moiseev-etal-2022-skill} approach. 
To conduct experiments using our AeroQA dataset, we compiled the Aviation corpus, comprising 665k lines of English text related to the aviation domain. This corpus was obtained by scraping 4,000 National Transportation Safety Board reports from the NTSB website, covering the period between 1981 and 2018. The reports, initially in PDF format, were converted to JSON format for easier processing. The paragraphs that contain clean text were extracted from selected sections of the reports which are Analysis, Probable Cause and Findings, and Factual Information. The selected paragraphs were then curated and included in the Aviation corpus, which served as a valuable resource for our research and experimentation.
\subsection{Experiment Data} \label{exp_data}


Our research utilizes the following datasets: 
a) Aviation Knowledge Graph (AviationKG) \cite{agarwal-etal-2022-knowledge} and AeroQA (Section \ref{aeroqa}): AviationKG is a knowledge graph designed explicitly for the aviation domain, while AeroQA is a question-answering dataset curated for multi-hop reasoning over AviationKG,
b) MetaQA \cite{zhang2018variational}: MetaQA consists of a knowledge base constructed from the WikiMovies dataset \cite{miller-etal-2016-key} and a set of question-answer pairs. It serves as a benchmark dataset for multi-hop reasoning. WikiMovies represents the movie domain. 
The statistics of these datasets are presented in Table \ref{tab:triplets_stats} and \ref{tab:QA_stats}.
\par
The QA datasets chosen for the experiments undergo preprocessing to make them suitable for the experimental procedures. During the preprocessing stage, if a single question has multiple answers separated by the `|' symbol, each answer is treated as a distinct instance of a question-answer pair. Rather than considering them as a single combined instance, they are split into individual question-answer pairs, with each answer associated with the same question. This separation facilitates improved handling and analysis of the data throughout the experiments.
\par
We chose these datasets deliberately because they cover diverse domains and exhibit variations in terms of size and characteristics, allowing us to evaluate the performance and generalizability of our proposed method across different contexts. This choice allows us to demonstrate the versatility of our approach across diverse datasets. 
 

 
\par
Additionally, Table \ref{tab:QA_stats} provides the statistics for the AviationQA dataset, which was utilized in the experimentation conducted by \citealt{agarwal-etal-2022-big}. However, we did not employ this dataset in our experiments due to its limitation in lacking multi-hop reasoning capabilities for QA tasks, as discussed in detail in Section \ref{aeroqa}.


\begin{table}[!h]
  \centering

  \captionsetup{position=below} 
  \setlength{\abovecaptionskip}{10pt}
  
  \begin{tabular}{ >{\centering\arraybackslash} p{3cm} >{\centering\arraybackslash} p{3cm}}
    \hline
    \textbf{Dataset} & \textbf{\# of triples} \\
    \hline
    AviationKG & 193,372 \\
    WikiMovies & 269,482 \\
    \hline
  \end{tabular}
  \caption{The statistics of triples (subject, relation, object) for two knowledge bases: AviationKG \cite{agarwal-etal-2022-knowledge} and WikiMovies \cite{zhang2018variational}.}
  \label{tab:triplets_stats}
\end{table}

\begin{table}[!h]
  \centering

  \captionsetup{position=below} 
  \setlength{\abovecaptionskip}{10pt}
  
  \begin{tabular}{ >{\centering\arraybackslash} p{2cm} >{\centering\arraybackslash} p{1.5cm} >{\centering\arraybackslash} p{1.5cm} >{\centering\arraybackslash} p{1.5cm}}
    \hline
    \textbf{Dataset} & \textbf{Train} & \textbf{Validation} & \textbf{Test} \\
    \hline
    MetaQA 1-hop & 96,106 & 9,992 & 9,947 \\
    MetaQA 2-hop & 118,980 & 14,872 & 14,872 \\
    MetaQA 3-hop & 114,196 & 14,274 & 14,274 \\
    \hline
    AeroQA 1-hop & 17,038 & 2,130 & 2,131 \\
    AeroQA 2-hop & 10,433 & 1,305 & 1,305 \\
    \hline
    AviationQA & 367,304 & 10,000 & 10,000 \\
    \hline
  \end{tabular}
  \caption{The statistics of question-answer pairs from the aviation and movie domains. The dataset MetaQA \cite{zhang2018variational} includes 1-hop, 2-hop, and 3-hop questions from the movies domain. AviationQA \cite{agarwal-etal-2022-big} specifically contains 1-hop questions from aviation domain. Our curated dataset, AeroQA, comprises both 1-hop and 2-hop questions. These statistics provide an overview of the question-answer distribution across different datasets used in our research.}
  \label{tab:QA_stats}
\end{table}


  

\section{Experimental Setup}


\begin{table*}[!ht]
\centering

  \captionsetup{position=below} 
  \setlength{\abovecaptionskip}{10pt}

\begin{tabular}{>{\centering\arraybackslash}m{6cm}>{\centering\arraybackslash}m{1.4cm}>{\centering\arraybackslash}m{1.4cm}>
{\centering\arraybackslash}m{1.4cm}>
{\centering\arraybackslash}m{1.4cm}>
{\centering\arraybackslash}m{1.4cm}}
\cline{1-6}
{\textbf{Models}} & \textbf{AeroQA 1-hop} & \textbf{AeroQA 2-hop} & \textbf{MetaQA 1-hop} & \textbf{MetaQA 2-hop} & \textbf{MetaQA 3-hop} \\
\cline{1-6}
T5-large (Baseline) & 52.88 & 41.57 & 24.5 & 32.65 & 42.31 \\
\cline{1-6}
T5-large + C4  & 51.38 & 40.65 & 23.53 & 32.78 & 39.66 \\
T5-large + Aviation\_Corpus & 52.04 & 41.73 & - & - & - \\
\cline{1-6}
T5-large + KG & 52.64  & 41.19 & 23.89 & 15.82 & 31.30 \\
\cline{1-6}
T5-large + C4 + KG (SKILL \cite{moiseev-etal-2022-skill}) & 54.66 & 41.34 & 71.47 & 33.57 & 43.41 \\
T5-large + Aviation\_Corpus + KG & 56.78 & 42.11 & - & - & - \\
\cline{1-6}
GPT-3 (Fine-tuned) & 24.30 & 20.99 & 18.73 & 16.71 & 54.77 \\
GPT-3.5-turbo (one-shot) & 0.37 & 2.22 & 53.90 & 21.07 & 23.06\\
\cline{1-6}
KITLM (Our method: T5-large + $relevant$ $KG-Triples$) & \textbf{86.06} & \textbf{43.52} & \textbf{91.26} & \textbf{71.19} & \textbf{71.62} \\
\cline{1-6}
\end{tabular}
\caption{
The table displays the exact match scores obtained on the test set for three models, T5-large, GPT-3, and GPT-3.5, utilized for the QA tasks. In our proposed KITLM approach (Section \ref{KITLM}), the term $relevant$ $KG-Triples$ represents distilled triples sourced from the knowledge base, which were utilized to provide contextual information for the questions. The table includes models labeled as T5-large + Z, which underwent continual pre-training with additional inputs (Z) such as text, KG triples, or a combination of text and triples. These pre-trained models were further fine-tuned for the QA task to generate exact match scores.
The performance of the KITLM approach is also compared with the state-of-the-art language models, namely GPT-3 and GPT-3.5-turbo. GPT-3 was fine-tuned on the corresponding dataset for the QA task.
The "-" symbol refers to non-applicability as the aviation corpus is a distinct domain and cannot be applied to the MetaQA dataset. In contrast, the C4 dataset, being a generic domain dataset,  applies to both MetaQA and AeroQA.
}
\label{tab:KG_DL_results}
\end{table*}



  


We applied the SKILL \cite{moiseev-etal-2022-skill} approach and our proposed KITLM approach on the T5-large model, which has 770M parameters. Additionally, we included the GPT-3 and GPT-3.5-turbo models for comparison with the knowledge-infused T5. 
\\\\
\noindent \textbf{SKILL.} The approach consists of two parts: continual pre-training and fine-tuning. 
In the process of continual pre-training, a balanced distribution is maintained by integrating both text and triples, ensuring an equal ratio of 50:50 between the two.
For T5-large, we conducted SKILL training for 20 epochs with a batch size of 32, followed by fine-tuning for 20 epochs with a batch size of 128. We used seeds 0 and 42 for continual pre-training and fine-tuning, respectively. 
During the training process, we utilized AdaFactor \cite{Shazeer2018AdafactorAL} as the optimizer with specific settings: a learning rate of 1e-3, scale\_parameter as False, relative\_step as False, and warmup\_init as False \footnote{\url{https://discuss.huggingface.co/t/t5-finetuning-tips/684/3}}. The maximum sequence length for both training and fine-tuning was set to 128, and a doc stride of 128 was applied during fine-tuning.
\\\\
\noindent \textbf{Baseline.} For baseline comparisons, we utilize pre-trained T5 checkpoints of the same size. In order to isolate the effect of knowledge infusion from the influence of additional text sources such as C4 and Aviation corpus used for pre-training, we follow a similar approach as the SKILL \cite{moiseev-etal-2022-skill}. However, since the code repository for the SKILL is not available, we implemented our own code for the method. 
To differentiate the effects of knowledge infusion, we create a second baseline by training the T5 checkpoints on the text for half of the previously mentioned steps. This adjustment ensures that the amount of text pre-training aligns with the SKILL model, allowing us to attribute any observed improvements to knowledge infusion. Furthermore, to assess the significance of text in conjunction with structured data, we create an additional T5 baseline that only utilizes triples. All other settings remain consistent with the SKILL. 
\\\\
\noindent \textbf{KITLM.} The approach comprises two main modules: (a) the retrieval module, which extracts triples from the knowledge base to provide contextual information, and (b) the fine-tuning module, which involves fine-tuning the T5-large model using the question+context combination. 
For single-hop QA, we retrieved the top 5 triples. In the first iteration of multi-hop QA, the top-k triples are retrieved with a value of k=3. In the subsequent iteration, k$^2$ triples are retrieved due to the reduction in triples after filtration (explained in Section \ref{KITLM}). This iteration process continues for N-hop QA, with N*k triples retrieved in the final iteration. For the fine-tuning of the model, a batch size of 128 is used, while the other settings are the same as the experimental setup of SKILL.
\\\\
\noindent \textbf{GPTs.} The GPT-3 and GPT-3.5-turbo models were accessed via the OpenAI API, and specifically, the GPT-3 model was fine-tuned on the QA tasks for AeroQA and MetaQA. During the fine-tuning of the GPT-3 model, we employed a batch size of 32 for AeroQA and 128 for MetaQA. To control the randomness of the model in training, a temperature of 0 was employed, and the model was trained for two epochs.
This decision was made based on the observation that the loss started converging by the second epoch for the Curie model.
To accommodate the multi-word factual answers present in the dataset, we have set the maximum token size to 50 for both GPT-3 and GPT-3.5-turbo.
For GPT-3.5-turbo, we utilized a prompt-based approach along with one-shot learning. The prompt instructed the model to predict the answer to a given question and to output "N/A" if the answer was not available. To construct the prompt, we included a random example from the development set of the corresponding datasets. However, specifically for GPT-3.5-turbo, we pre-processed the dataset by removing the square brackets from both the test set and the example included in the prompt. This adjustment was implemented because it was observed that removing the brackets slightly improved the performance compared to when the brackets were not removed.
\\\\
\noindent \textbf{Evaluation.}
During the evaluation, as part of pre-processing, both the correct answers and predicted answers of the test set for all models are converted to lowercase.
The exact match score is utilized as the evaluation metric on the test set for all experiments  In the case of a QA dataset where a question can have multiple correct answers, the scoring is determined based on the following criteria: If the predicted answer matches any of the correct answers associated with a question, a score of 1 is assigned. Conversely, if the predicted answer does not match any of the correct answers, a score of 0 is assigned. 
Unlike the T5 model, which predicts only one answer and matches it with the gold answers, GPTs are generative models that generate answers. We determine if any of the gold answers are present in the generated answer and assign a value of 1 if there is a match and 0 otherwise. 

\section{Results and Analysis} \label{result}


In table \ref{tab:KG_DL_results}, we depict the results for the question-answering task and compare the performance of our proposed framework, KITLM, and different knowledge integration settings using the T5-large model. The evaluation is conducted on the AviationQA and MetaQA datasets. Additionally, the table includes the performance of GPT-3 and GPT-3.5-turbo (ChatGPT).
\par
The T5-large model is used as the baseline, while the state-of-the-art knowledge infusion method, described in \cite{moiseev-etal-2022-skill}, is referred to as the SKILL pre-trained combined model. This model combines T5 with additional input, which is unstructured corpora denoted as X, and also incorporates triples as part of the knowledge infusion process.
The T5 + X + KG demonstrates improved performance compared to T5, T5 + X, and T5 + KG for both MetaQA and AeroQA datasets. 
In MetaQA, the performance is adversely affected when using only non-verbalized KG-Triples, leading to a decrease in scores compared to the C4 text experiment. 
Although the T5 + KG-Triples model exhibits a slight improvement in the AeroQA, it fails to surpass the T5 baseline due to catastrophic interference, despite its inherent domain-specific benefits.
The T5 + KG-Triples result underscores the importance of incorporating unstructured corpora for language models, as relying solely on triples can result in catastrophic forgetting. 
But this doesn't mean that the integration of triples is irrelevant, and infusing only text into the language model will help.
We observed a decline in the performance for the tasks after utilization of only C4 text compared to the baseline. 
The rationale behind the ongoing use of text-only pre-training is explained in \cite{raffel2020exploring}. It is proposed that repeatedly training on C4 text could potentially lead to a decrease in performance for a T5 model.
Although incorporating the aviation corpus in the AeroQA task significantly improves the score compared to the C4 text, this improvement can be primarily attributed to the domain similarity between the corpus and the task itself.
To conclude, considering the individual inclusion of unstructured text and triples, the outcomes vary, with some cases showing a decline in performance while others exhibit slight improvements. However, when both text and triples are combined using the SKILL, the results demonstrate enhancements. It is important to acknowledge that explaining the performance of these approaches can be complex, and their effectiveness relies on the particular corpora employed, such as C4 or the Aviation Corpus. Furthermore, the continual pre-training demands higher computational resources.


\begin{table*}
    \centering
      \captionsetup{position=below} 
  \setlength{\abovecaptionskip}{10pt}
    \begin{tabular}{>{\centering\arraybackslash} >{\centering\arraybackslash}p{2cm} >{\centering\arraybackslash}p{5cm} >{\centering\arraybackslash}p{2.5cm} >{\centering\arraybackslash}p{5cm} >{\centering\arraybackslash}p{1.5cm}}
        \hline
        \textbf{Dataset} & \textbf{Question} & \textbf{Gold Answer} & \textbf{ChatGPT (prompt)} & \textbf{KITLM} \\
        \hline
        AeroQA 1-hop & What environmental issue caused
[AccidentNumber\_IAD05LA071]? & Tailwheel & N/A (The given Accident Number does not provide any
information related to an environmental issue) & tailwheel \\
\hline
        AeroQA 2-hop & What is the aircraft category of the registered aircraft involved in [AccidentNumber\_LAX04LA084]? & Airplane & turbo-jet-turbofan-turboprop-turboshaft & airplane \\
        \hline
        
        MetaQA 1-hop & what movies were [Jessica Simpson] an actor in? & Employee of the Month|Blonde Ambition & Employee of the Month| Blonde Ambition|Private Valentine: Blonde \& Dangerous|The Love Guru & employee of the month \\
        \hline
        
        MetaQA 2-hop & What are the primary languages in the movies directed
by [David Mandel]? & German & N/A & german \\
\hline

        MetaQA 3-hop & The movies that share actors with the movie [Waxworks] were in which languages? & Swedish|German| French|English & N/A & german  \\
        \hline
    \end{tabular}
    \caption{The table compares examples from the AeroQA and MetaQA datasets, showcasing the differences between ChatGPT and our proposed framework, KITLM. }
    \label{tab:mytable}
\end{table*}

\par
To address the aforementioned limitations, we developed the KITLM approach, demonstrating the best performance across all tasks. 
The reason why KITLM performs better is as it selects the most relevant triples from the knowledge base after mitigating noise. The triples serve as the context for answering the question. 
KITLM is a selective approach that helps to remove confusing elements compared to existing methods, such as SKILL.  
This resulted in KITLM to achieve a significant improvement of 30\% and 20\% in the exact match score for the AeroQA 1-hop and MetaQA 1-hop tasks, respectively. 
Attributing to similar reasons, KITLM demonstrates better performance for the 2-hop and 3-hop QA. 
The reason KITLM surpasses other methods is due to its adeptness in handling multi-hop question-answering tasks, which demand advanced reasoning abilities for accurate answers. Unlike previous infusion methods that relied on continual pre-training, which can result in catastrophic forgetting, KITLM addresses this challenge effectively by removing the pre-training procedure to infuse knowledge. 
Another advantage of KITLM is its ability to mitigate the computational power requirements associated with knowledge infusion methods, which typically arise from pre-training.
KITLM addresses this issue by selecting triples N times for N-hop QA, eliminating the need to train the model repeatedly. This approach not only reduces computational costs but also improves performance.
The significance of the KITLM is amplified by its capability to adapt to diverse domains and tasks, eliminating the necessity for domain-specific corpora in the process of knowledge infusion.
\\
\\
\textbf{Performance on GPT-3 and ChatGPT.} 
In our experiments on MetaQA and AeroQA datasets, we observed that ChatGPT, despite being a powerful model, encountered challenges in producing accurate results.
In table \ref{tab:mytable}, we depict some examples from both datasets.
ChatGPT is able to provide some answers for 1-hop MetaQA questions. This  can be attributed to the fact that ChatGPT has movie domain knowledge. However, ChatGPT face challenge is 2-hop and 3-hop questions. 
In the case of AeroQA, ChatGPT faces challenges even in the 1-hop question. This can be because chatGPT might lack knowledge of a specialized domain, aviation.  On the other hand, KITLM is able to answer all the questions in table \ref{tab:mytable}.
%
These observations highlight the limitations of ChatGPT in domain-specific question-answering tasks and the effectiveness of the KITLM in achieving higher accuracy with relevant knowledge infusion. 
It becomes evident that smaller language models, when combined with knowledge infusion techniques, can achieve better accuracy than LLMs for the QA task.
\par
Following the evaluation of ChatGPT's performance with one-shot learning, we fine-tune the GPT-3 model for the QA tasks.
However, the results obtained for AeroQA 1-hop and MetaQA 1-hop were relatively poor, with EM scores of 24.3\% and 18.73\% respectively. Also, the 2-hop and 3-hop tasks performed worse than KITLM. Due to space limitation, we provide a comprehensive result of ChatGPT and GPT-3 on the entire MetaQA and AeroQA datasets at an anonymous link \footnote{\url{https://anonymous.4open.science/r/KITLM_CIKM23-8BDF/}}.

%
%
\par
Our results using GPT-3 and ChatGPT highlight the significant advantage of knowledge infusion for smaller models compared to large language models such as the GPT series. Through the incorporation of knowledge infusion techniques, we achieved superior performance across various tasks, demonstrating the effectiveness of leveraging domain-specific knowledge to enhance the capabilities of smaller models.








\section{Conclusion and Future Work}
We introduce a novel framework called KITLM, which addresses the challenge of providing context for multi-hop question answering. KITLM leverages a knowledge base by filtering and selecting relevant information to derive the necessary context for accurately answering the question.
Significantly, our study highlights that even with the advent of large language models such as GPT-3 and ChatGPT, utilizing a knowledge base remains crucial for accomplishing domain-specific tasks. This emphasizes the ongoing importance of incorporating domain-specific knowledge and context in language models to enhance their performance and effectiveness in specialized domains.
We have successfully developed a dataset called AeroQA that caters explicitly to multi-hop question-answering tasks in the aviation domain. This dataset is valuable for tasks involving reasoning and complex queries within the aviation domain. Furthermore, we have contributed an Aviation corpus, which is a useful resource for knowledge infusion tasks in language models (LMs).
\par
In future work, there is a focus on harnessing the reasoning capabilities of knowledge bases to enhance tasks such as sentiment analysis and summarization, thereby improving the performance and contextual understanding of these NLP tasks. However, it is important to acknowledge the limitations of existing knowledge sources, such as knowledge graphs, and the abundance of available language models. Consequently, future efforts will explore ways to integrate domain-specific knowledge into language models, potentially replacing traditional knowledge bases. This integration aims to optimize the utilization of knowledge in a more efficient and effective manner.

\begin{acks}
To Robert, for the bagels and explaining CMYK and color spaces.
\end{acks}

\bibliographystyle{ACM-Reference-Format}
\bibliography{sample-base}

\appendix









\end{document}